\documentclass[lettersize,journal]{IEEEtran}
\usepackage{amsmath,amssymb,amsfonts}
\usepackage{algorithmic}
\usepackage{algorithm}
\usepackage{array}
\usepackage[caption=false,font=normalsize,labelfont=sf,textfont=sf]{subfig}
\usepackage{textcomp}
\usepackage{stfloats}
\usepackage{url}
\usepackage{xurl}
\usepackage{verbatim}
\usepackage{graphicx}
\usepackage{cite}
\usepackage{makecell}
\usepackage{booktabs}
\usepackage[table]{xcolor}
\usepackage{booktabs}
\usepackage{array}
\usepackage{multirow}
\definecolor{blockA}{RGB}{245,248,255}    
\definecolor{blockB}{RGB}{248,245,255}    
\definecolor{highlight}{RGB}{235,245,235} 
\definecolor{best}{RGB}{220,240,255}     
\definecolor{good}{RGB}{235,245,255}     
\definecolor{bad}{RGB}{255,235,235}  
\definecolor{blockA}{RGB}{235,242,252}    
\definecolor{blockB}{RGB}{252,243,233}    
\definecolor{highlight}{RGB}{232,246,236} 
\definecolor{human}{RGB}{240,240,248}     
\begin{document}

\title{UniPPTBench: A Unified Benchmark for Presentation Generation Across Diverse Input Settings}

\author{Bo Zhao, Maosheng Pang, Chen Zhang, Huan Yang, Yixin Cao, Wei JI}

\author{Bo Zhao, Maosheng Pang, Chen Zhang, Huan Yang, Yixin Cao, Wei Ji
\thanks{Bo Zhao, Maosheng Pang and Chen Zhang contribute equally to this project.}
\thanks{Corresponding author: Wei Ji, weiji@nju.edu.cn}
}


\maketitle

\begin{abstract}
Recent advances in multimodal generative models have made automatic presentation generation increasingly feasible. However, existing research and systems remain limited in both scenario coverage and evaluation specificity. Prior work typically studies presentation generation under isolated input settings, whereas real-world use cases span diverse scenarios, including vague user prompts, long documents, multimodal materials, and multiple heterogeneous sources. Moreover, current evaluations are often insufficiently scenario-specific. They mainly rely on generic presentation-quality criteria, such as visual appeal, layout quality, and overall coherence, but fail to assess the core capabilities required by different input settings, including grounded compression, visual-text alignment, and cross-source synthesis. Consequently, the field lacks a unified benchmark and a scenario-aware evaluation framework for faithfully diagnosing presentation-generation systems across diverse real-world settings.
We present UniPPTBench, a unified benchmark for presentation generation across four representative input settings: vague-prompt, long-document, multimodal-document, and multi-source generation. We further introduce UniPPTEval, a scenario-aware evaluation protocol that combines shared metrics for cross-setting comparison with scenario-specific metrics tailored to each setting’s core requirements. We also provide transparent reference baselines to support reproducible comparison.
Experiments on UniPPTBench reveal substantial performance variation across settings and recurring failure modes in content grounding, multimodal integration, and cross-source synthesis. In particular, strong performance on generic presentation-quality metrics does not necessarily imply strong task fulfillment in grounded scenarios. Together, UniPPTBench and UniPPTEval provide a faithful and diagnostic foundation for evaluating presentation generation across diverse real-world scenarios.
Code and data will be publicly available.
\end{abstract}


\begin{figure*}[!t]
\centering
\includegraphics[width=\textwidth]{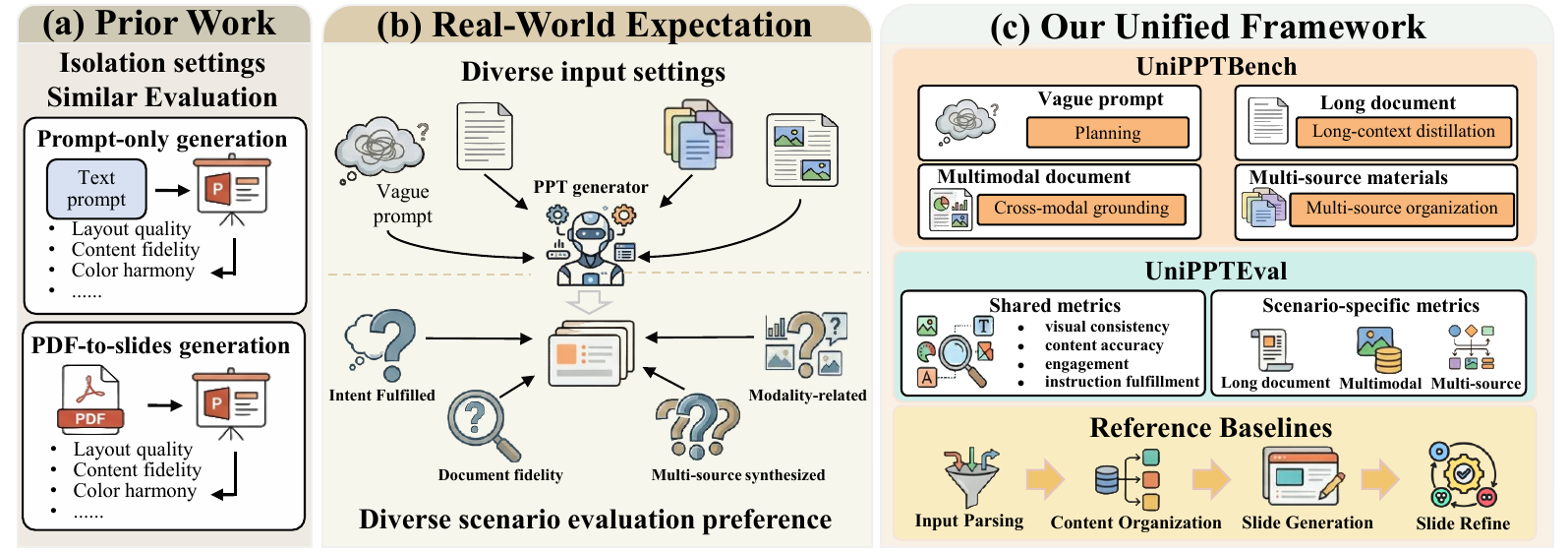}
\caption{Motivation and overview of our framework. (a) Prior work usually studies presentation generation in isolated settings, such as prompt-only or PDF-to-slides generation, with largely similar generic metrics. (b) Real-world PPT generation involves more diverse inputs and evaluation needs, including vague prompts, long documents, multimodal files, and multi-source materials. (c) We therefore propose UniPPTBench for unified benchmarking across input settings, and UniPPTEval for combining shared and scenario-specific evaluation.}

\label{fig:teaser}
\end{figure*}
\section{Introduction}

Recent advances in multimodal generative models~\cite{gpt5,yin2024mllm} have made automatic presentation generation increasingly feasible. Given a high-level request or a collection of source materials, modern systems can now produce slide decks with structured content, visual elements, and coherent layouts. This capability has broad potential in education, business communication, research reporting, and content creation. Unlike conventional text generation, however, presentation generation is a visually grounded communication task: a system must generate fluent textual content, organize information into spatially structured slides, align claims with visual evidence, and maintain a coherent narrative across pages.


A key challenge in real-world presentation generation is the diversity of input scenarios. Users may start from vague intent, long textual documents, multimodal materials, or multiple heterogeneous sources. Although these scenarios share the same output format—a complete slide deck—they impose different core requirements, ranging from open-ended planning and faithful compression to visual grounding and cross-source synthesis.
Despite rapid progress, existing research has not yet provided a unified way to study these input regimes. Recent academic works have explored multimodal large language models and agent-based frameworks~\cite{react,wang2024agents,xie2024lma} for presentation generation from user instructions or source documents. However, most open-source systems are designed for a single input setting, such as instruction-only or document-conditioned generation~\cite{ge2025autopresent,fu2022doc2ppt,zheng2025pptagent,yang2025autoslides,liang2025slidegen}. As a result, it remains unclear whether a method that performs well in one setting can generalize to others, especially when the demand shifts from open-ended planning to grounded compression, multimodal evidence alignment, or multi-source synthesis. Meanwhile, commercial systems have started to support end-to-end slide generation from prompts or external materials~\cite{kimi_slides_2025,gemini_canvas_2025,skywork_slides_2025,notebooklm_slides_2025,manus_slides_2025}, but their internal workflows, training data, and evaluation procedures are usually not publicly disclosed, limiting transparent comparison and systematic diagnosis.

This lack of unified task coverage also makes evaluation more difficult. Existing assessments often focus on generic presentation-quality dimensions, such as visual appeal, layout quality, readability, and content coherence~\cite{ge2025autopresent,zheng2025pptagent,chen2026presentbench}. While necessary, these dimensions are insufficient for judging whether a system fulfills the task imposed by a specific input scenario. A deck may look polished and well organized while omitting key evidence from a source document, misrepresenting a chart, or failing to reconcile duplicated or conflicting information across sources. In other words, visual quality and task fulfillment are related but not equivalent. Without scenario-aware evaluation, current benchmarks may overestimate presentation-generation ability by rewarding outputs that resemble good slides but are weakly grounded in the input~\cite{faitheval,factscore}. These limitations call for a controlled evaluation infrastructure that can diagnose both general deck quality and the input-conditioned capabilities required by different presentation-generation scenarios.

To address these challenges, we introduce UniPPTBench, a unified benchmark for presentation generation across four representative input settings: vague-prompt, long-document, multimodal-document, and multi-source generation. These settings are organized along a progression of grounding complexity. Vague-prompt generation evaluates open-ended planning under underspecified user intent. Long-document generation tests source-faithful compression and content selection. Multimodal-document generation emphasizes alignment between textual claims and visual materials such as figures, charts, and tables. Multi-source generation further requires evidence integration, redundancy removal, and faithful synthesis across heterogeneous sources. By covering these settings under a common task formulation, UniPPTBench enables systematic comparison across real-world workflows.

Building on UniPPTBench, we propose UniPPTEval, a scenario-aware evaluation protocol that explicitly separates general presentation quality from input-conditioned task fulfillment. UniPPTEval combines shared metrics for deck-level quality, including structure, coherence, readability, and visual organization, with scenario-specific metrics targeting the core capability required by each setting. For example, long-document generation is evaluated for source coverage and faithful compression; multimodal-document generation for visual-text grounding and evidence alignment; and multi-source generation for cross-source integration, deduplication, and synthesis. This design reveals failure modes hidden by generic presentation-quality metrics alone.

Moreover, a unified agent-based baseline for presentation generation is developed. Our UniPPTAgent is among the first open-source baselines designed to support multiple presentation-generation input settings within a unified framework. Unlike prior open-source systems tailored to one input regime~\cite{ge2025autopresent,fu2022doc2ppt,zheng2025pptagent,yang2025autoslides,liang2025slidegen}, our baseline uses a shared generation interface while adapting its planning, grounding, and slide-composition modules to different input types, which achieves state-of-the-art performance among open-source methods, providing a strong and reproducible reference point for future research rather than a minimal pipeline.

Experiments on UniPPTBench reveal several important findings. First, system performance varies substantially across input settings, confirming that single-scenario evaluation gives an incomplete estimate of presentation-generation capability. Second, generic presentation-quality metrics do not reliably reflect grounded task fulfillment: many systems generate visually plausible and coherent decks while failing to preserve key source content, align claims with visual evidence, or synthesize relations across documents. Third, multimodal-document and multi-source settings remain especially challenging, exposing limitations in evidence grounding, information integration, deduplication, and faithful synthesis. These findings suggest that presentation generation should be evaluated not only by whether the output looks like a high-quality deck, but also by whether it satisfies the scenario-specific requirements imposed by its input.

Our contributions are summarized as follows:

1. We introduce UniPPTBench, a unified benchmark for presentation generation covering four representative real-world input settings: vague-prompt, long-document, multimodal-document, and multi-source generation.

2. We propose UniPPTEval, a scenario-aware evaluation protocol that combines shared presentation-quality metrics with setting-specific metrics for input-conditioned task.

3. We develop a transparent open-source reference baseline that supports all four input settings and provides a strong reproducible comparison point.

4. We conduct extensive experiments and diagnostic analysis, revealing persistent limitations of current systems in grounded compression, multimodal alignment, cross-source integration, and faithful synthesis.

\section{Related Work}

Recent advances in large language models and multimodal foundation models have made automatic presentation generation increasingly feasible~\cite{yang2025autoslides,2024enhancing,zheng2026deeppresenter,liang2025slidegen,xu2025pregenie,kumar2026learning}. Existing work has mainly developed along two task formulations: instruction-driven generation and document-conditioned generation.

Instruction-driven methods generate slides directly from natural-language requests. AutoPresent~\cite{ge2025autopresent}, for example, formulates NL-to-slide generation and introduces SlidesBench for instruction-conditioned evaluation. Document-conditioned methods instead take a source document, typically a scientific paper or report, as input. DOC2PPT~\cite{fu2022doc2ppt} introduced document-to-slide generation with paired paper--slide data, while PPTAgent~\cite{zheng2025pptagent} further studies document-based generation and evaluates outputs along content, design, and coherence dimensions.

These works have advanced both system development and early evaluation for slide generation, but their formulations remain closely tied to specific input assumptions. Instruction-oriented benchmarks focus on prompts or design requests, whereas document-oriented systems usually assume a single structured source. Their evaluations are correspondingly built around reference similarity, holistic quality assessment, or coarse multi-dimensional judgments. SlidesBench combines reference-based and reference-free evaluation for instruction-conditioned generation; PPTEval~\cite{zheng2025pptagent} uses instance-agnostic MLLM-based scoring across broad dimensions; and PresentBench~\cite{chen2026presentbench} improves human alignment through rubric-based, instance-specific assessment.

However, real-world presentation creation rarely starts from only a short instruction or a single clean document. Users often prepare slides from underspecified intent, long reports, multimodal materials, or multiple heterogeneous sources that must be selected, reconciled, and organized into a coherent narrative. This gap motivates a unified benchmark that explicitly spans diverse input settings, together with an evaluation framework that combines cross-setting comparability with scenario-specific capability assessment.

\section{UniPPTBench} \label{sec:pptbench} 
To bridge academic settings and practical applications, we introduce UniPPTBench, a benchmark for evaluating presentation generation across diverse, realistic input conditions. Unlike prior benchmarks focused on a single input paradigm~\cite{chen2026presentbench}, UniPPTBench systematically covers four representative scenarios: \textbf{Vague-Prompt} for open-ended content planning, \textbf{Long-Document} for faithful compression, \textbf{Multi-Modal} for joint visual-text reasoning, and \textbf{Multi-Source} for cross-document synthesis~\cite{ge2025autopresent,fu2022doc2ppt,zheng2025pptagent}. All settings output standard slide decks but isolate distinct input-conditioned capabilities.

\subsection{Construction Pipeline} 

The construction of UniPPTBench consists of four stages: \textit{source collection}, \textit{task formulation}, \textit{metadata annotation}, and \textit{quality control}.

To ensure open-source compliance and data diversity, source materials are gathered from authoritative public sources: business financial reports from SEC EDGAR, academic papers from arXiv, policy reports from the OECD and WEF, international affairs materials from the UN/UNCTAD, and educational content from MIT OpenCourseWare and document repositories like Theseus.

During the task formulation stage, three researchers paired the source documents with realistic user intents. Specifically, 48 intents were independently authored by humans, while the remaining intents were initially drafted by Gemini~3 Flash~\cite{gemini3} using few-shot examples and subsequently revised by the authors to ensure consistency with the source materials and target task settings.

For metadata annotation, we constructed structured reference annotations for grounding-intensive scenarios, including manually identified key points and their corresponding evidence spans in the source materials, which are further utilized in the scenario-specific evaluation protocol described in Sec.~\ref{sec:ppteval}.

To ensure instance validity, we further conducted manual pass/fail quality control on all instances. Specifically, we verified whether (1) the user intent was supported by the accompanying source materials, (2) the input structure conformed to the designated scenario setting, and (3) the task could be completed without relying on inaccessible external evidence. Instances failing any criterion were revised or removed from the benchmark.

Overall, this construction pipeline preserves the diversity and realism of presentation generation tasks while providing structured and reproducible support for model evaluation.

\subsection{Dataset Characteristics} 
The main statistics of UniPPTBench are summarized in Table~\ref{tab:dataset_stats}. The benchmark comprises 126 tasks across four input settings and six application domains. We deliberately skew the data distribution toward complex, visually grounded scenarios: \textit{multimodal-document} constitutes the largest split (38.1\%), compelling models to move beyond pure text processing and perform active visual grounding. 

Furthermore, the scale of the input materials rigorously tests models' long-context processing capabilities. The source materials for \textit{long-document} tasks average 127.78 pages, while \textit{multi-source} tasks reach an average total length of 59.39 pages, supported by an average of 4.5 distinct documents per task. This forces models to execute dense information compression and resolve cross-document redundancy, rather than relying on naive text concatenation. 

Domain-wise, the dataset is predominantly anchored in Research \& Academic (34.1\%) and Business \& Marketing (24.6\%). This aligns with common high-value use cases and ensures that the generated slides are rigorously evaluated on both knowledge-intensive synthesis and professional formatting standards.

\begin{table}[t]
\centering
\small
\setlength{\tabcolsep}{6pt}
\renewcommand{\arraystretch}{1.15}
\caption{Main statistics of UniPPTBench.}
\label{tab:dataset_stats}

\begin{tabular}{lccc}
\toprule
\textbf{Category} & \textbf{\#} & \textbf{\%} & \textbf{Avg. Pages} \\
\midrule

\multicolumn{4}{l}{\textbf{Task Setting}} \\
Vague-Prompt        & 27 & 21.4 & 0 \\
Long-Document       & 32 & 25.4 & 127.78 \\
Multimodal-Document & 48 & 38.1 & 25.50 \\
Multi-Source        & 19 & 15.1 & 59.39 \\

\addlinespace[4pt]

\multicolumn{4}{l}{\textbf{Domain}} \\
Research \& Academic     & 43 & 34.1 & \\
Business \& Marketing   & 31 & 24.6 & \\
Policy \& Governance    & 18 & 14.3 & \\
Creative \& Design      & 14 & 11.1 & \\
Education \& Tutorial   & 12 & 9.5  & \\
Technology \& Product   & 8  & 6.3  & \\

\bottomrule
\end{tabular}
\end{table}
\begin{figure*}[t]
    \centering
    \includegraphics[width=\textwidth]{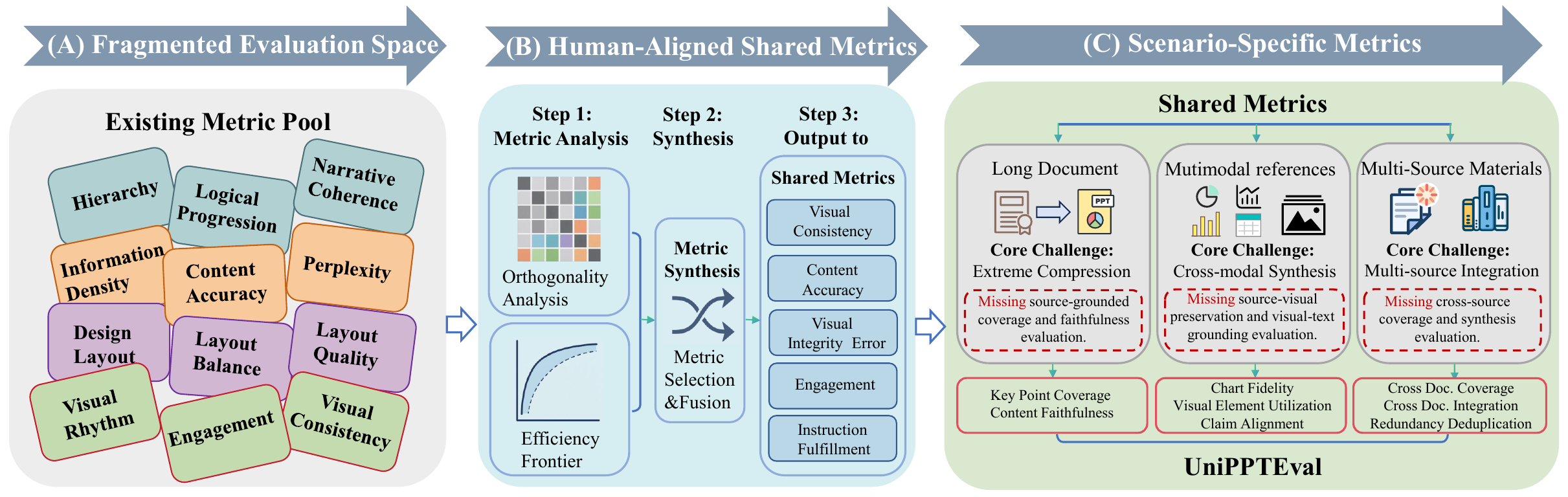}
    \caption{Overview of UniPPTEval. Candidate shared metrics are normalized, aligned with human preferences, filtered through orthogonality and efficiency considerations, and reduced to a compact shared metric set. This shared set is complemented by annotation-driven scenario-specific metrics for grounding-intensive settings.}
    \label{fig:ppteval_pipeline}
\end{figure*}

\section{UniPPTEval}
\label{sec:ppteval}

Evaluating presentation generation across diverse input settings introduces a fundamental tension: the protocol must provide a common basis for cross-setting comparison while maintaining strict sensitivity to scenario-specific constraints. Existing approaches often fail to balance these needs, relying instead on loosely defined and entangled criteria~\cite{zheng2025pptagent,chen2026presentbench}. This ambiguity not only limits diagnostic interpretability but also degrades alignment with human judgment. To resolve this, UniPPTEval introduces a factorized evaluation design driven by a principled metric construction pipeline (illustrated in Fig.~\ref{fig:ppteval_pipeline}). This section details the derivation of shared metrics, the necessity and design of scenario-specific metrics, and the validation of the overall protocol.

\subsection{Principled Metric Construction}

Rather than assuming a fixed set of evaluation dimensions, UniPPTEval derives shared metrics from a comprehensive candidate pool through a sequence of steps: \textit{candidate collection}, \textit{human alignment}, \textit{orthogonality analysis}, and \textit{efficiency--accuracy frontier selection}.

Metric candidates are collected from prior work on presentation generation and slide evaluation, including PPTAgent~\cite{zheng2025pptagent}, AutoPresent~\cite{ge2025autopresent}, PresentBench~\cite{chen2026presentbench}, and related systems. The resulting pool spans multiple dimensions: content correctness, instruction fulfillment, structural organization, visual quality, and presentation-level effectiveness.

To align with human judgment, candidate metrics are evaluated against human preference signals, following recent frameworks~\cite{faitheval,factscore}. Metrics with weak predictive utility or unstable behavior are discarded. The remaining metrics undergo orthogonality analysis to remove redundancy and retain complementary signals.

Building on this reduced candidate space, UniPPTEval selects a compact shared metric subset along an efficiency--accuracy frontier. This establishes a minimal yet reliable configuration that maintains strong human alignment while avoiding redundancy and evaluation overhead. 
The final shared metrics are as follows: Instruction Fulfillment, Engagement, Content Accuracy, Visual Consistency, and Visual Integrity. These metrics jointly capture task-level alignment, communicative effectiveness, factual reliability, and visual quality dimensions of generated presentations. Details are provided in the appendix.

Together, these metrics define a unified evaluation space that supports direct comparison across heterogeneous input settings while maintaining semantic consistency.

\begin{figure}[t]
    \centering
    \includegraphics[width=\linewidth]{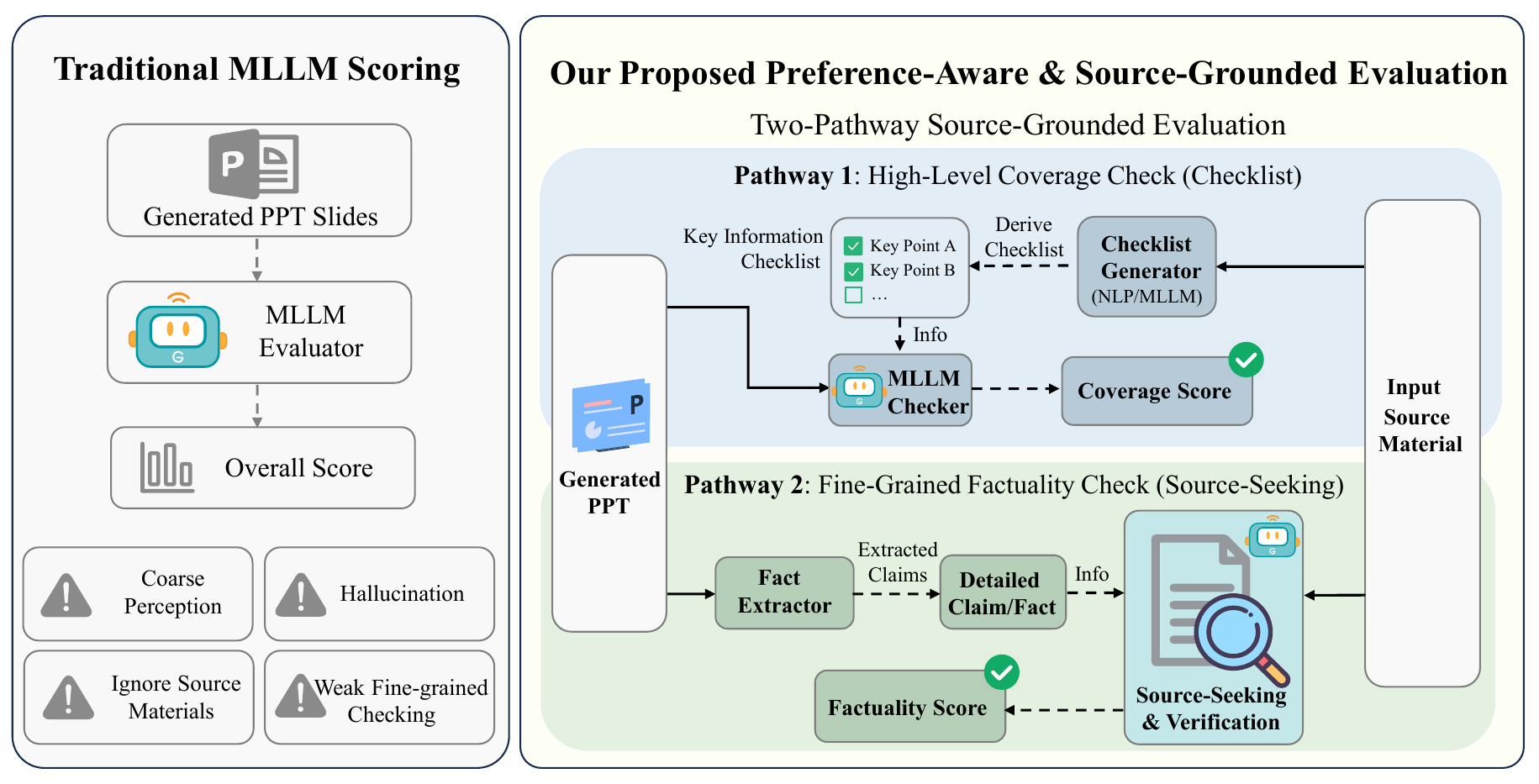} 
    \caption{Comparison between traditional MLLM scoring (left) and our proposed Source-Grounded Evaluation (right). Traditional methods rely on black-box end-to-end perception and are prone to evaluation hallucinations. In contrast, UniPPTEval adopts a two-pathway framework: Checklist-based coverage evaluation (Pathway 1) and Source-Seeking factual verification (Pathway 2), enabling more reliable and faithful assessment.}
    \label{fig:contrast}
\end{figure}

\subsection{Scenario-Specific Metrics}

In practical presentation generation tasks, visual appeal does not necessarily guarantee task fulfillment. Generated slides may still suffer from deep-seated issues such as critical information omission, chart misrepresentation, or unresolved cross-document redundancies. Our controlled perturbation experiments in Section~\ref{sec:validation} and Table~\ref{tab:manus_perturbation_validity} provide empirical support for this phenomenon: generic Shared Metrics are often insensitive to these source-conditioned generation failures, whereas scenario-specific metrics can precisely identify the same perturbations. These experimental results confirm that while Shared Metrics provide a foundation for quality evaluation, they fall short in diagnosing whether a model faithfully adheres to source material constraints. Furthermore, given the massive and heterogeneous nature of real-world inputs (such as long documents or multi-source packets), directly relying on Large Language Models (LLMs) for end-to-end comparison is highly prone to evaluation hallucinations and the omission of critical details~\cite{ji2023survey,liu2024lost}.

To bridge this gap, UniPPTEval introduces the Source‑Grounded Evaluation paradigm, shifting the benchmark from visual perception to rigorous evidence tracing. As shown on the right of Fig.~\ref{fig:contrast}, this is implemented via a two‑pathway verification logic that decouples global evaluation into two dimensions of evidence comparison:

\begin{itemize}
\item \textbf{Macro-level Coverage Check (Pathway 1):} The core philosophy is to avoid pre-defining subjective standards of quality, and instead construct a key information \textit{Checklist} based on the source material. By verifying the extent to which the generated slides cover the key points in the checklist, it ensures task completeness at a macro level.
\item \textbf{Micro-level Factuality Pursuit (Pathway 2):} The core philosophy is to require that every factual claim in the slides must be evidence-based. Through a \textit{Source-Seeking} mechanism, the PPT content is decomposed into specific statement items and verified one by one against the source materials for truthfulness, thereby producing quantitative scores with high diagnostic value.
\end{itemize}

This factorized evaluation design, which uses source materials as the yardstick, effectively transforms a vague overall impression into deterministic local judgments, enhancing the transparency and interpretability of the evaluation.

We provide a complete list of scenario-specific metrics for reproducibility. Due to space constraints, we formally elaborate representative metrics that capture the core evaluation principles in each setting, while only brief descriptions are provided for the remaining metrics. \textbf{The full definitions are provided in the appendix.} We will open-source the evaluation code later.

\paragraph{\textit{Long-Doc}}
The long-document setting focuses on source-grounded compression and content selection from extended materials. We define two metrics:

\begin{itemize}
\item \textbf{Key Point Coverage}
This metric measures whether the generated slides cover annotated source-constrained content units, such as required objective facts, arguments, concepts, or section-level takeaways. Each coverage point is independently judged based on semantic equivalence rather than surface matching, and the final score is aggregated over all annotated items.

\item \textbf{Content Faithfulness}
Unlike coverage evaluating recall, faithfulness assesses the factual precision of the generated content. This metric verifies factual accuracy by extracting statements from the slides and validating them against the input source materials~\cite{kryscinski2020evaluating,factscore}.
\end{itemize}

\paragraph{\textit{Multi-Modal}}
The multimodal setting evaluates the system's ability to invoke visual evidence (such as charts, illustrations, etc.), with its essence lying in \textit{visual grounding}. For each task, annotations identify critical visual elements and additional relevant information. We define three metrics:

\begin{itemize}
\item \textbf{Visual Element Utilization}
This metric evaluates whether the model leverages visual evidence from the input rather than relying solely on textual restatement.

\item \textbf{Figure--Claim Alignment}
This is a deep visual grounding metric. Drawing on the philosophy of Pathway 2, it verifies whether textual claims are substantively supported by corresponding visual evidence.

\item \textbf{Chart Translation Fidelity}
This metric captures distortions or hallucinations introduced during chart redrawing or transformation. The focus is on preserving chart semantics rather than pixel-level visual replication~\cite{chartqa}.
\end{itemize}

\paragraph{\textit{Multi-Source}}
The multi-source setting evaluates the model’s ability to synthesize information across multiple heterogeneous input sources. For each task, annotations specify source-specific contributions, cross-source integration requirements, and overlap groups. We define three metrics:

\begin{itemize}
\item \textbf{Cross-Document Coverage}
This metric measures whether the generated slides adequately cover key information identified from each source in the annotations.

\item \textbf{Cross-Document Integration}
This metric evaluates whether information from multiple sources is synthesized into a coherent and unified narrative, penalizing outputs that only present isolated summaries without establishing cross-source semantic connections.

\item \textbf{Redundancy Deduplication}
This metric measures whether redundant information across sources is consolidated into a unified expression. The evaluation is based on annotated overlap groups, which identify content likely to be repeated under naive source concatenation. This metric captures the model’s ability to detect and remove redundancy during multi-source generation.
\end{itemize}

\paragraph{\textit{Vague-Prompt}} This setting is evaluated only with shared metrics due to the lack of source-grounded references.

\subsection{Validation of the Evaluation Protocol}

We validate UniPPTEval from three complementary perspectives: validity, reliability, and robustness. Specifically, we verify that the proposed metrics respond to capability-relevant perturbations, exhibit stable behavior under repeated evaluation, and remain consistent across different judge models. These analyses ensure that the evaluation protocol provides reliable and interpretable signals for benchmark comparison. Detailed validation settings and empirical results are presented in Section~\ref{sec:experiments} .

\section{UniPPTAgent Framework}
\begin{figure*}[t] 
    \centering 

    \includegraphics[width=\textwidth]{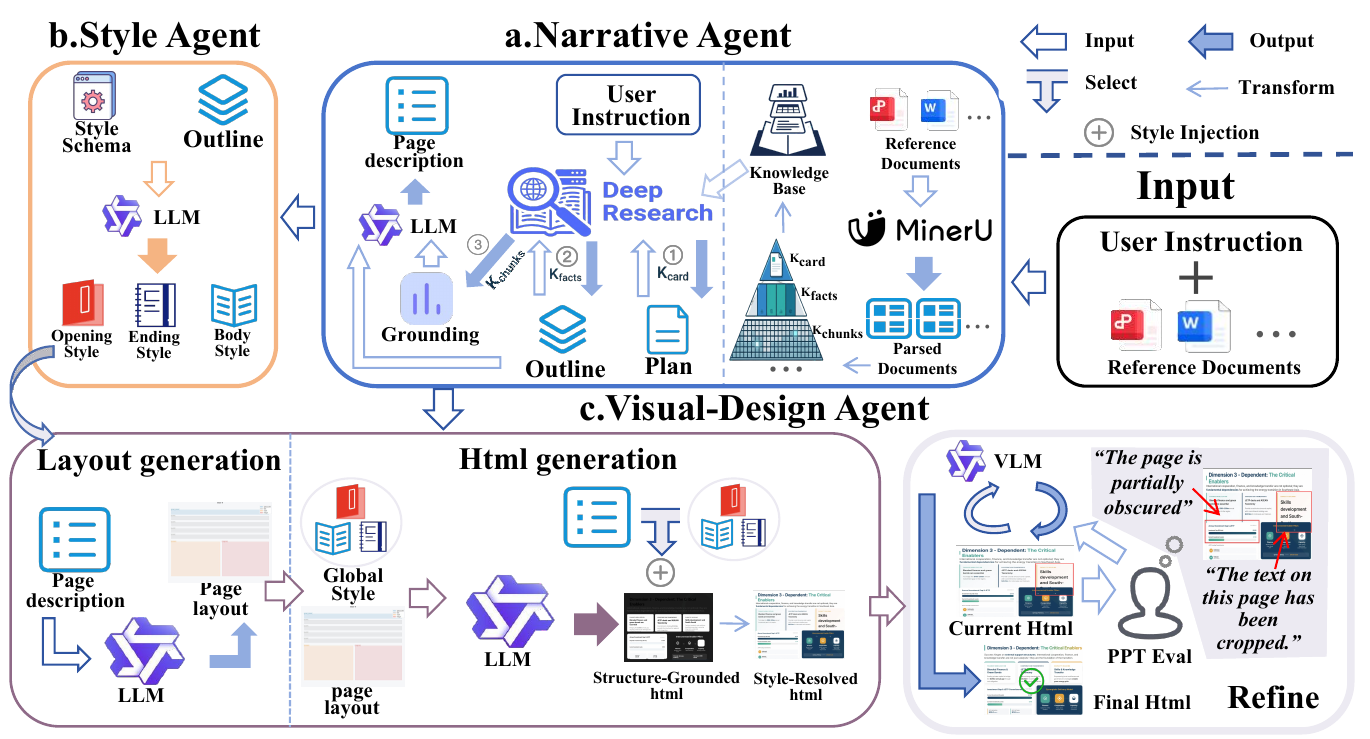}
    \caption{\textbf{Overall architecture of the proposed multi-agent pipeline.} The framework consists of three core components: (1) the Narrative Agent, which performs deep research to transform user instructions and reference documents into structured outlines and page descriptions; (2) the Style Agent, which induces a document-level style contract from a parameterized schema; and (3) the Visual Design Agent, which translates layouts into style-resolved HTML through an iterative refinement loop.}
    \label{fig:framework} 
\end{figure*}
We formulate presentation generation as the task of mapping natural-language instructions and/or reference documents to a coherent, multi-page slide deck.
Following recent industrial practice, we adopt a web-native slide representation in which each slide is encoded directly in HTML/CSS~\cite{manus_slides_2025}.
Formally, our framework $F$ is defined as
\begin{equation}
  F \colon \bigl([\textit{Ins}],\; [\textit{Doc}_1, \dots, \textit{Doc}_n]\bigr)
    \;\longrightarrow\; P,
  \quad P = \{H_i\},
\label{eq:struct}
\end{equation}
where $\textit{Ins}$ denotes an optional user instruction, $\textit{Doc}_1, \dots, \textit{Doc}_n$ denotes an optional set of reference documents, $P$ denotes the complete presentation, and $H_i$ denotes the HTML slide for each page.
This unified formulation accommodates diverse input needs and formats within a single processing pipeline.
The goal of UniPPTAgent is not to propose a new generation model, but to provide a transparent multi-setting reference pipeline for controlled benchmarking. Its design contribution lies in unified input normalization, scenario-aware grounding, and reusable evaluation-compatible intermediate representations, which make heterogeneous presentation-generation settings comparable within one framework.

The same input can yield many equally valid presentations, differing in narrative structure, visual layout, and stylistic choices. These concerns are not only heterogeneous but also interdependent, making it difficult to handle them reliably in a single pass. We therefore decompose the task into three specialized agents (Fig.~\ref{fig:framework}), each targeting a distinct stage of the generation process:
the \textbf{Narrative Agent} normalizes heterogeneous inputs and progressively structures them into slide outlines and page-level descriptions;
the \textbf{Style Agent} induces a document-level style contract to ensure cross-page visual consistency;
and the \textbf{Visual Design Agent} renders layout- and style-grounded HTML slides, correcting perceptual defects via an iterative feedback loop.
This decomposition keeps each concern independently addressable, facilitating both modular development and targeted capability diagnosis.

\subsection{Narrative Agent: From Inputs to Outline and Page Descriptions}
\label{sec:narrative_agent}

Presentation generation involves fundamentally different challenges depending on what the input provides: instructions alone lack factual grounding, documents alone lack narrative direction, and multiple documents additionally require cross-source synthesis rather than naive concatenation~\cite{deyoung2024multidoc}.
A single processing step is insufficient to handle this diversity reliably.
To address this, we adopt a progressive strategy that first normalizes heterogeneous inputs into a uniform representation and then structures them stage by stage into two intermediate representations: a multi-page outline $O$ and page descriptions $D$.
Formally, the Narrative Agent pipeline is defined as
\begin{equation}
  \text{NarrativeAgent} \colon
    \bigl([\textit{Ins}],\,[\textit{Doc}_1, \dots, \textit{Doc}_n]\bigr)
    \,\rightarrow\,(O,\, D),
\label{eq:narrative_agent}
\end{equation}
This process comprises two phases: \textbf{input normalization} and \textbf{progressive content structuring}.

\paragraph{Input Normalization.}
To provide downstream stages with a consistent input interface regardless of the original input type, we normalize all inputs into a unified $(K, \textit{Ins})$ representation, where $K$ is a structured knowledge base extracted from the documents and $\textit{Ins}$ captures the user's instruction.
When documents are provided, each document undergoes independent structured parsing to construct three granularities of knowledge (corresponding to the hierarchical knowledge base in Fig.~\ref{fig:framework}~(a)): a condensed overview (\textit{card}) for high-level planning, section-level summaries (\textit{facts}) for outline generation, and fine-grained text passages (\textit{chunks}) for evidence retrieval. These three representations are merged into a unified knowledge base $K$, while user instructions serve directly as $\textit{Ins}$. When only instructions are given, $K = \varnothing$; when only documents are given, $\textit{Ins}$ is empty. This normalization ensures that all subsequent stages receive a consistent input interface.

\paragraph{Progressive Content Structuring.}
As illustrated by the data flow in Fig.~\ref{fig:framework}~(a), we generate $(O, D)$ through a progressive \textit{plan--outline--ground} procedure~\cite{wei2022cot}, drawing on progressively finer granularities of knowledge from $K$ at each stage---coarser representations guide high-level planning while finer ones support detailed grounding---each step operates within a manageable scope while maximizing content fidelity:
\begin{equation}
\begin{gathered}
  (K_{\text{card}},\, \textit{Ins})
    \xrightarrow{\;\text{Narrative Planning}\;} S \\
  \xrightarrow[\scriptstyle K_{\text{facts}}]{\;\text{Outline Induction}\;} O
    \xrightarrow[\scriptstyle K_{\text{chunks}}]{\;\text{Evidence Retrieval}\;} G
\end{gathered}
\label{eq:deepresearch}
\end{equation}
Specifically: (i)~\textbf{Narrative Planning} determines the global narrative structure, the number of slides, and each slide's role based on the condensed overview $K_{\text{card}}$ and $\textit{Ins}$, producing a slide plan $S$;
(ii)~\textbf{Outline Induction}, guided by $S$, extracts relevant section-level summaries $K_{\text{facts}}$ and expands them into a detailed outline $O$ containing per-page titles, key messages, and content points;
(iii)~\textbf{Evidence Retrieval} retrieves concrete supporting evidence for each page from fine-grained passages $K_{\text{chunks}}$ according to $O$, using a source-aware retrieval strategy~\cite{gao2024rag} that prioritizes content from relevant documents, yielding page-level grounding $G$.
Finally, the outline $O$ and grounding $G$ are fused by an LLM to generate page descriptions $D=\{d_i\}$, providing structured input for downstream visual generation.

\subsection{Style Agent: Document-Level Style Induction}
\label{sec:style_agent}

Generating a coherent multi-page presentation requires that typography, colors, and layout conventions remain consistent across all slides. If style decisions are deferred to page-level generation, inconsistencies accumulate and the deck loses its document-level identity. We therefore pre-compute a unified global style before any page is rendered. To avoid the tension between unconstrained generation and rigid templates, we introduce a Style Schema that reframes style induction as constrained slot filling over a set of predefined parameterized slots.
Formally, the style induction process is defined as
\begin{equation}
  \text{StyleAgent} \colon (O,\; S_{\text{schema}})
    \;\longrightarrow\; S_{\text{global}},
\label{eq:style_agent}
\end{equation}
where $O$ is the narrative outline, $S_{\text{schema}}$ is the Style Schema, and $S_{\text{global}} = \{S_{\text{opening}},\, S_{\text{body}},\, S_{\text{ending}}\}$ is the resulting global style. During HTML generation, pages \textit{apply} role-appropriate style modules rather than regenerating styles, enforcing consistency through systematic style binding.

\paragraph{Style Schema.}
The Style Schema structures style induction into explicit slots---color tokens, typographic hierarchies, spacing scales, and component primitives---defining a constrained design space that ensures consistency while permitting parameterized variation.

\paragraph{Modular Decomposition}
Beyond enforcing a consistent vocabulary of style tokens, the Style Agent further recognizes that a uniform style cannot accommodate the functional differences across page roles: opening slides need to capture attention, body slides focus on content delivery, and ending slides provide closure. The Style Agent decomposes $S_{\text{global}}$ into three role-specific modules ($S_{\text{opening}}$, $S_{\text{body}}$, $S_{\text{ending}}$, which share the same foundational style system while adapting to their respective communicative functions. This structured style representation yields explicit, verifiable targets that support fine-grained diagnosis of cross-page style consistency.

\subsection{Visual Design Agent: Perceptual Grounding and Iterative Correction}
\label{sec:visual_design_agent}

Visual rendering presents two distinct sources of failure: flawed spatial organization and low-level rendering artifacts. We address both through a staged pipeline that explicitly separates layout planning, HTML generation, and perceptual refinement into sequential, independently addressable steps.
Formally, the visual generation pipeline is defined as
\begin{equation}
  \text{VisualDesignAgent} \colon (D,\; S_{\text{global}})
    \;\longrightarrow\; P = \{H_i\},
\label{eq:visual_agent}
\end{equation}
where $D = \{d_i\}$ denotes the page descriptions from the Narrative Agent, $S_{\text{global}}$ denotes the global style from the Style Agent, and $P$ denotes the final presentation. As illustrated in Fig.~\ref{fig:framework}~(c), we decompose generation into three phases: (i)~\textit{layout planning}~\cite{inoue2023layoutdm,hsu2023posterlayout}, where an LLM constructs an abstract spatial blueprint from each page description $d_i$ to stabilize downstream generation; (ii)~\textit{HTML generation}, which produces structured HTML under layout and style constraints; and (iii)~\textit{perceptual refinement}, where a feedback loop diagnoses and corrects residual visual defects.

\paragraph{HTML Generation.}
Given the layout blueprint and the global style $S_{\text{global}}$, the LLM produces structure-grounded HTML for each page. To enforce cross-page stylistic consistency, the HTML is instantiated with the appropriate role-specific style module ($S_{\text{opening}}$, $S_{\text{body}}$, $S_{\text{ending}}$), yielding style-resolved HTML. This style injection ensures consistency by applying a fixed document-level contract rather than regenerating styles for each page.

\paragraph{Perceptual Refinement.}
Despite producing structurally valid HTML, the LLM may still introduce residual perceptual defects that are difficult to anticipate through generation alone. We therefore introduce a closed-loop refinement stage in which each rendered page is inspected by an automated evaluation module that diagnoses perceptual violations and assesses visual appeal~\cite{nima,hou2025tip,celona2022tip,zeng2020tip,li2020tip}. The resulting diagnostics---together with the rendered screenshot and current HTML---are passed to a VLM, which generates a targeted patch to correct the flagged defects while enhancing overall aesthetic quality. This loop iterates until visual quality converges or a preset iteration limit is reached~\cite{madaan2023selfrefine}.

\section{Experiments}
\label{sec:experiments}

\begin{figure*}[t]
    \centering
    \includegraphics[width=0.95\textwidth]{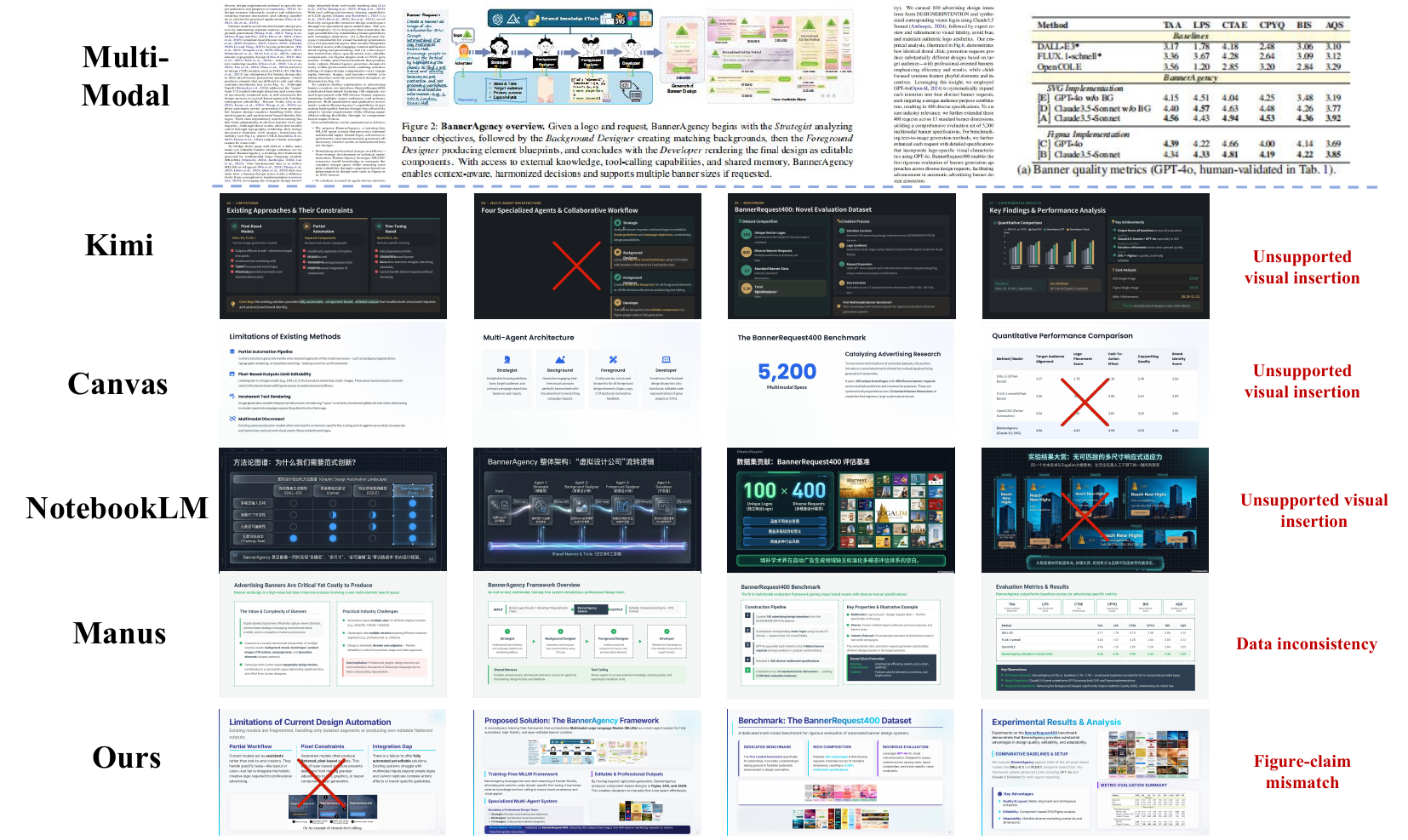}
    \caption{Qualitative comparison of representative PPT generation systems under a Multi-Modal input setting. Despite using the same input materials and instructions, different systems exhibit distinct failure modes while also sharing certain common errors. }
    \label{fig:error1}
\end{figure*}

\begin{figure*}[t]
    \centering
    \includegraphics[width=0.95\textwidth]{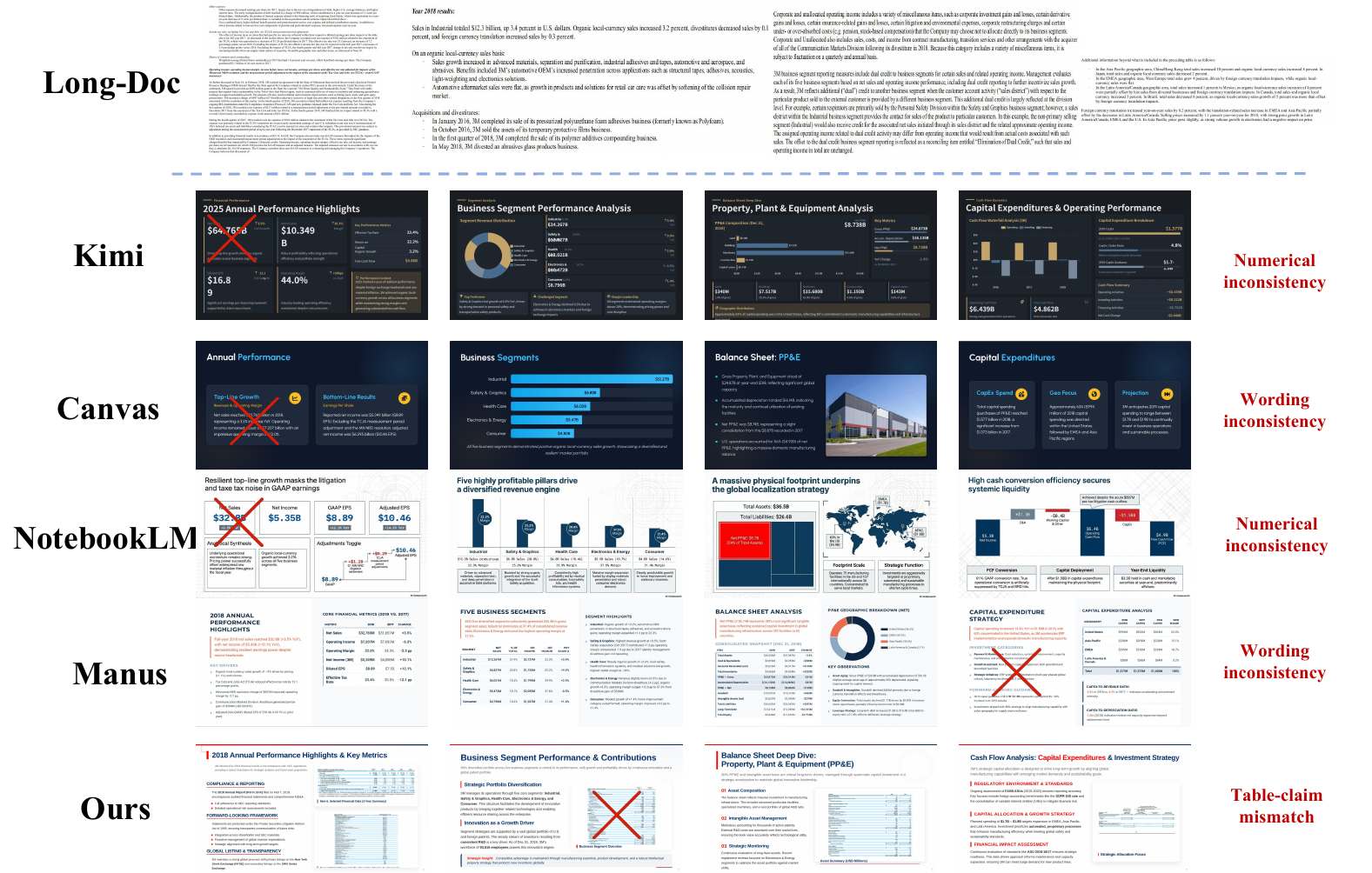}
    \caption{Qualitative comparison of representative PPT generation systems under the Long-Doc input setting. It can be observed that most errors primarily arise from factual errors or hallucinations, as well as improper use of source content. }
    \label{fig:error2}
\end{figure*}

This section evaluates automatic presentation generation systems on UniPPTBench using UniPPTEval. We describe the experimental setup, present comparative results and key findings, examine UniPPTEval’s validity through validation experiments, and ablate UniPPTAgent’s components.

\subsection{Experimental Setup}
\label{sec:exp_setup}

\paragraph{Compared systems.} The evaluation includes a diverse set of current presentation generation systems: \textsc{NotebookLM}~\cite{notebooklm_slides_2025}, \textsc{Manus}~\cite{manus_slides_2025}, \textsc{Canvas}~\cite{gemini_canvas_2025}, \textsc{Kimi}~\cite{kimi_slides_2025}, \textsc{Skywork}~\cite{skywork_slides_2025}, \textsc{AutoSlides}~\cite{yang2025autoslides}, \textsc{AutoPresent}~\cite{ge2025autopresent}, \textsc{EvoPresent}~\cite{evopresent}, and two internal variants, \textsc{Ours-Gemini} and \textsc{Ours-Qwen}. For all systems, we provide the same user intent and background materials, and request the generation of a complete slide deck. Each method is used according to its default or officially recommended configuration.
Specifically, \textsc{Manus} is used in the slide generation mode (Manus~1.6), \textsc{Canvas} is instantiated via the Gemini~3 Flash Canvas interface, \textsc{Kimi} is used in its Agent PPT mode with intelligent layout enabled, and \textsc{Skywork} is used in its PPT generation mode. For API-based methods, \textsc{AutoSlides} and \textsc{EvoPresent} are both implemented using Gemini~3 Flash as the backbone model, while \textsc{AutoPresent} is implemented with Gemini~3 Flash under a high-level instruction setting with creative mode enabled.

For systems that output PPTX or PDF files, results are uniformly converted into ordered slide images before evaluation, ensuring a consistent input format for the judge. 

\paragraph{Implementation details of our baselines.}
Both \textsc{Ours-Gemini} and \textsc{Ours-Qwen} instantiate the UniPPTAgent framework with the same three-agent architecture---Narrative Agent, Style Agent, and Visual Design Agent---and differ only in the backbone model used for style induction, page-description synthesis, layout planning, HTML generation, and iterative repair: Gemini~3 Flash~\cite{gemini3} and Qwen~3.6~\cite{qwen3.6}, respectively.

For document-conditioned settings, reference materials are first parsed with MinerU~\cite{mineru} and normalized into the hierarchical knowledge representation described in Sec.~\ref{sec:narrative_agent}. The Narrative Agent then follows the plan--outline--ground procedure: DeepSeek-V3~\cite{deepseek_v3} produces the narrative plan and outline, retrieves page-level textual evidence from the normalized knowledge base, and passes the outline and retrieved evidence to the variant backbone to synthesize structured page descriptions. For multimodal inputs, extracted visual assets are matched to relevant pages through a coarse-to-fine visual evidence alignment procedure and appended to the corresponding page descriptions.

The Style Agent uses the variant backbone to fill the Style Schema and produce a document-level global style. The Visual Design Agent then generates layout blueprints and style-grounded HTML slides, followed by a closed-loop perceptual refinement stage that renders each page, detects visual defects, and repairs the HTML until convergence or a preset iteration limit. Detailed preprocessing parameters, retrieval settings, visual-alignment thresholds, decoding configurations, and refinement criteria are provided in appendix

\paragraph{Benchmark and protocol.} Experiments are conducted on UniPPTBench (Sec.~\ref{sec:pptbench}), which covers four representative input settings. We follow the UniPPTEval framework (Sec.~\ref{sec:ppteval}) for automatic evaluation, using Gemini~3 Flash~\cite{gemini3} as the default judge. To improve stability, metrics are computed via separate rubric-guided calls and averaged over three runs. For methods that support only a subset of settings (e.g., \textsc{AutoPresent}, \textsc{AutoSlides}, \textsc{EvoPresent}), results are reported only on the tasks they can complete.

\paragraph{Reported scores.}
In all tables, \textbf{Shared} is the equally weighted mean of five shared metrics. \textbf{Avg} is the scenario-level full score, averaged over all metrics within each setting.

For \emph{Vague-Prompt}, no scenario-specific metrics are defined; therefore, \textbf{Avg} is identical to \textbf{Shared}. For other settings, additional scenario-specific metrics are reported to capture task-dependent capabilities: \emph{Long-Doc} includes Key Coverage and Faithfulness, \emph{Multi-Source} includes Source Coverage , Integration , and Deduplication , and \emph{Multi-Modal} includes Visual Use (Vis.\ Use), Figure--Claim Alignment (Fig.-Claim), and Chart Fidelity (Chart Fid.). The \textbf{Avg} score aggregates both shared and scenario-specific metrics for each setting.

In addition, we report \textbf{Human Rank} obtained from user studies, together with the Spearman correlation ($\rho$) between automatic scores and human preferences for each setting, to assess alignment with human judgment.

As an external sanity check, we include a subset of human-authored slide decks as \textsc{Human-Reference}. This is not a directly comparable system, but verifies that the evaluation framework assigns appropriately high scores to high-quality human presentations.

\begin{table*}[t]
\centering
\scriptsize
\setlength{\tabcolsep}{3.6pt}
\renewcommand{\arraystretch}{1.08}
\caption{Scenario-level results for \textit{Vague-Prompt} and \textit{Multi-Source}. Human Rank is obtained from user studies, and $\rho$ measures its correlation with model scores. Colors indicate method groups: light blue for closed-source methods, light orange for open-source methods, and light green for our methods. "-"indicates that the corresponding method cannot generate slides under the given input scenario.}
\label{tab:scenario_detail_vague_multisource}
\resizebox{\textwidth}{!}{
\begin{tabular}{lccc cccccc}
\toprule
Method
& \multicolumn{3}{c}{Vague-Prompt}
& \multicolumn{6}{c}{Multi-Source} \\
\cmidrule(lr){2-4} \cmidrule(lr){5-10}
& Shared & Avg & \makecell{Human Rank \\ {\scriptsize ($\rho=0.52$)}}
& Shared & Source Cov. & Integration & Deduplication & Avg & \makecell{Human Rank \\ {\scriptsize ($\rho=0.89$)}} \\
\midrule

\rowcolor{blockA}
NotebookLM
& 8.47 & 8.47 & 1
& 8.44 & 4.29 & 5.22 & 10.00 & 7.71 & 2 \\

\rowcolor{blockA}
Manus
& 9.40 & 9.40 & 2
& 8.35 & 7.79 & 7.78 & 7.92 & 8.16 & 1 \\

\rowcolor{blockA}
Canvas
& 8.90 & 8.90 & 6
& 7.14 & 6.38 & 3.70 & 7.40 & 6.65 & 7 \\

\rowcolor{blockA}
Kimi
& 4.56 & 4.56 & 4
& 5.07 & 7.30 & 6.89 & 10.00 & 6.19 & 6 \\

\rowcolor{blockA}
Skywork
& 8.40 & 8.40 & 5
& 6.48 & 6.37 & 4.93 & 8.33 & 6.50 & 5 \\

\rowcolor{blockB}
AutoSlides
& -- & -- & --
& -- & -- & -- & -- & -- & -- \\

\rowcolor{blockB}
AutoPresent
& 4.72 & 4.72 & 8
& -- & -- & -- & -- & -- & -- \\

\rowcolor{blockB}
EvoPresent
& -- & -- & --
& -- & -- & -- & -- & -- & -- \\

\rowcolor{highlight}
\textbf{Ours-Gemini}
& 8.43 & 8.43 & 3
& 6.87 & 7.92 & 7.76 & 9.38 & 7.43 & 3 \\

\rowcolor{highlight}
\textbf{Ours-Qwen}
& 7.85 & 7.85 & 7
& 6.92 & 7.60 & 8.04 & 7.88 & 7.26 & 4 \\
\midrule
Human-Reference
& 8.30 & 8.30 & --
& 8.92 & 7.15 & 7.51 & 10.00 & 8.66 & -- \\

\bottomrule
\end{tabular}
}
\end{table*}

\begin{table*}[t]
\centering
\scriptsize
\setlength{\tabcolsep}{3.2pt}
\renewcommand{\arraystretch}{1.08}
\caption{Scenario-level results for \textit{Long-Doc} and \textit{Multi-Modal}. Human Rank is obtained from user studies, and $\rho$ measures its correlation with model scores. Colors indicate method groups: light blue for closed-source methods, light orange for open-source methods, and light green for our methods. "-"indicates that the corresponding method cannot generate slides under the given input scenario.}
\label{tab:scenario_detail_longdoc_multimodal}
\resizebox{\textwidth}{!}{
\begin{tabular}{lccccc cccccc}
\toprule
Method
& \multicolumn{5}{c}{Long-Doc}
& \multicolumn{6}{c}{Multi-Modal} \\
\cmidrule(lr){2-6} \cmidrule(lr){7-12}
& Shared & Key Cov. & Faith. & Avg & \makecell{Human Rank \\ {\scriptsize ($\rho=0.72$)}}
& Shared & Vis. Use & Fig.-Claim & Chart Fid. & Avg & \makecell{Human Rank \\ {\scriptsize ($\rho=0.92$)}} \\
\midrule

\rowcolor{blockA}
NotebookLM
& 8.33 & 7.47 & 3.88 & 7.57 & 2
& 9.60 & 8.83 & 8.83 & 6.00 & 9.02 & 1 \\

\rowcolor{blockA}
Manus
& 9.49 & 9.55 & 6.26 & 9.04 & 1
& 9.13 & 4.33 & 3.87 & 8.40 & 7.78 & 2 \\

\rowcolor{blockA}
Canvas
& 8.81 & 8.55 & 5.39 & 8.28 & 6
& 7.67 & 1.50 & 1.92 & 6.20 & 6.00 & 8 \\

\rowcolor{blockA}
Kimi
& 6.96 & 9.48 & 8.28 & 7.50 & 4
& 5.26 & 4.17 & 3.12 & 8.00 & 5.20 & 9 \\

\rowcolor{blockA}
Skywork
& 7.13 & 5.11 & 3.83 & 6.37 & 3
& 8.84 & 3.23 & 2.17 & 8.00 & 7.20 & 6 \\

\rowcolor{blockB}
AutoSlides
& 4.78 & 4.90 & 3.93 & 4.67 & 8
& 7.06 & 9.50 & 9.50 & 8.75 & 7.88 & 3 \\

\rowcolor{blockB}
AutoPresent
& -- & -- & -- & -- & --
& -- & -- & -- & -- & -- & -- \\

\rowcolor{blockB}
EvoPresent
& 5.23 & 4.74 & 1.48 & 4.63 & 9
& 7.66 & 7.93 & 7.93 & 6.60 & 7.60 & 5 \\

\rowcolor{highlight}
\textbf{Ours-Gemini}
& 6.70 & 7.40 & 5.01 & 6.71 & 5
& 8.14 & 8.03 & 8.97 & 8.00 & 8.25 & 4 \\

\rowcolor{highlight}
\textbf{Ours-Qwen}
& 6.79 & 6.96 & 5.00 & 6.56 & 7
& 7.41 & 8.33 & 7.37 & 7.40 & 7.52 & 7 \\
\midrule
Human-Reference
& 9.20 & 9.39 & 10.00 & 9.34 & --
& 8.74 & 8.08 & 8.85 & 8.50 & 8.64 & -- \\

\bottomrule
\end{tabular}
}
\end{table*}

\subsection{Main Results}
\label{sec:main_results}
Table~\ref{tab:scenario_detail_vague_multisource} and Table~\ref{tab:scenario_detail_longdoc_multimodal} report the results across all four input settings. Overall, current automatic PPT generation systems already exhibit clear performance variations, but still fall significantly short of human-created presentations in reliably handling source-grounded content and producing high-quality slides in realistic scenarios.

Across generation methods, \textsc{Manus} and \textsc{NotebookLM} achieve the strongest overall performance among automated systems and remain competitive across different input settings. \textsc{Human-Reference} obtains consistently high scores, especially in source-grounded scenarios, suggesting that UniPPTEval can distinguish high-quality human-authored presentations from automated outputs.

\paragraph{Alignment with human preference.}
Across all input settings, automatic evaluation correlates positively with human judgment. In particular, the correlation is substantially stronger in settings with scenario-specific metrics, while relatively lower in \emph{Vague-Prompt} ($\rho=0.52$). This suggests that incorporating input-dependent evaluation criteria leads to better alignment with human perception of presentation quality.

We summarize the main findings below.

\paragraph{Finding 1: Grounding capability remains the primary bottleneck for reliable generation}

Across all evaluated tasks, systems consistently achieve higher scores on generic shared metrics than on scenario-specific dimensions, revealing a significant gap between visual fluency and actual task fulfillment.While visually coherent slide generation has shown significant improvement in recent years, its fidelity and alignment with source-grounded content remain unstable. In multimodal scenarios, for instance, systems frequently fail at figure--claim alignment or struggle to utilize key visual elements effectively, as illustrated in Fig.~\ref{fig:error1}. This often results in "hallucinated" layouts that rely on generic templates rather than the provided visual evidence. Similarly, in \emph{Long-Doc} tasks, content faithfulness is consistently lower than coverage, indicating that while models can extract key points, they often fail to reflect the source material accurately, leading to factual distortions in the compressed output as shown in Fig.~\ref{fig:error2}. Consequently, future improvements must transition from prioritizing aesthetics to developing stronger grounding mechanisms, such as precise source tracking and evidence preservation.

\paragraph{Finding 2: Cross-source synthesis poses a higher-order challenge than content extraction}

Even when key information is successfully retrieved, systems frequently struggle to organize it into a coherent, non-redundant narrative. This is particularly evident in \emph{Multi-Source} scenarios, where integration and deduplication scores remain unstable despite moderate shared scores. Unlike human-authored presentations—which excel at establishing semantic connections and narrative flow—automated systems often achieve higher coverage than integration. They tend to present isolated summaries of individual documents rather than a unified narrative that resolves cross-document redundancies. This contrast suggests that multi-input presentation generation is not merely a retrieval problem, but a high-level synthesis challenge. It requires explicit modeling of cross-source relationships to transform heterogeneous fragments into a structured, professional, and logically synthesized flow.

\paragraph{Finding 3: support for diverse input settings remains fragmented in the open-source ecosystem}
While open and reproducible systems show promising progress, their support across diverse input settings remains uneven. Several systems lack coverage for grounded or multi-input scenarios, and even when supported, performance tends to degrade more sharply than in proprietary systems. This indicates that robust multi-input presentation generation remains only partially addressed in the current open-source ecosystem, both in terms of capability and system coverage.
\subsection{Human Preference Study}

To complement automatic evaluation, we conduct a human preference study to assess overall presentation quality from a user-centric perspective. For each input setting, we randomly sample 5 cases and present annotators with outputs from three method per task. Annotators are asked to rank the presentations based on overall quality.

We adopt a relative ranking protocol, where annotators assign 3, 2, and 1 points to the first-, second-, and third-ranked presentations, respectively. Final method-level scores are computed by averaging these point assignments across cases and annotators, from which rankings are derived.

The study involves three annotators with substantial PPT design experience. Inter-annotator agreement measured using intraclass correlation coefficient (ICC) ranges from 71.3\% to 80.2\% across input settings, indicating substantial consistency.

\subsection{Validation of the Evaluation Protocol}
\label{sec:validation}

We validate UniPPTEval from three complementary perspectives: validity, reliability, and robustness.

\paragraph{Validity: Controlled Perturbation.}
To assess whether scenario-specific metrics capture intended capabilities rather than generic presentation quality, we introduce controlled perturbations at the output level by localized modifications to generated slide decks, systematically weakening specific capability dimensions. Specifically, for \emph{Long-Doc}, we delete or corrupt key information segments to disrupt content completeness and factual consistency; for \emph{Multi-Modal}, we modify or replace visual elements to break figure--claim alignment and degrade chart fidelity; and for \emph{Multi-Source}, we remove partial source content, weaken cross-document integration, or introduce redundancy to impair cross-source synthesis.

Table~\ref{tab:manus_perturbation_validity} presents representative results under these output-level perturbations. We observe that performance degradation is concentrated on the perturbed capability dimensions, while shared metrics exhibit comparatively smaller changes. This pattern indicates that UniPPTEval responds selectively to capability-specific degradation rather than merely reflecting overall presentation quality.

\begin{table}[t]
\centering
\small
\setlength{\tabcolsep}{4pt}
\renewcommand{\arraystretch}{1.05}
\caption{Controlled-perturbation results on \texttt{manus}. We report relative score changes only. Shared is the mean of Instruction Fulfillment, Engagement, Content Accuracy, Visual Consistency, and Visual Integrity. $\Delta\%=\frac{\text{Pert.}-\text{Orig.}}{\text{Orig.}}\times 100\%$. }
\label{tab:manus_perturbation_validity}
\begin{tabular}{lc}
\toprule
Metric & $\Delta\%$ \\
\midrule
\multicolumn{2}{l}{\textbf{Long-Doc}} \\
Shared Avg. & -5.85\% \\
Key Coverage & -12.05\% \\
Faithfulness & -12.82\% \\
\midrule
\multicolumn{2}{l}{\textbf{Multi-Source}} \\
Shared Avg. & -0.23\% \\
Source Coverage & -11.75\% \\
Integration & -4.10\% \\
Redundancy (Dedup) & -12.00\% \\
\midrule
\multicolumn{2}{l}{\textbf{Multi-Modal}} \\
Shared Avg. & -5.76\% \\
Visual Utilization & -43.22\% \\
Figure--Claim Alignment & -35.77\% \\
Chart Fidelity & -21.62\% \\
\bottomrule
\end{tabular}
\end{table}

\paragraph{Reliability.}
We assess evaluation stability by repeating scoring under identical tasks, annotations, and prompt configurations. As summarized in Table~\ref{tab:reliability_robustness}, standard deviations of the aggregate score (\textit{Avg}) remain low across input settings, ranging from 0.11 to 0.18. This indicates UniPPTEval yields stable scores under repeated LLM-based evaluation.

\paragraph{Robustness.}
We further evaluate robustness by replacing the default judge with a GPT-based evaluator and comparing the resulting system rankings and score shifts. Table~\ref{tab:reliability_robustness} shows that Spearman correlations between Gemini- and GPT-based rankings exceed 0.9 in all settings, indicating highly consistent system-level conclusions across evaluators. The mean score differences between judges are also small, with limited variance across settings, suggesting minimal evaluator-specific bias. These results show that UniPPTEval provides stable and judge-invariant evaluation signals, which we attribute to its rubric-based and factorized design that reduces sensitivity to evaluator-specific preferences.

\begin{table}[t]
\centering
\small
\setlength{\tabcolsep}{3.5pt}
\renewcommand{\arraystretch}{1.05}
\caption{
Reliability and robustness of UniPPTEval across input settings. 
Repeat Std. denotes the standard deviation of the aggregate \textit{Avg} score over three repeated runs. 
$\rho$ denotes the Spearman rank correlation between Gemini- and GPT-based evaluators, and $\Delta$ denotes GPT minus Gemini.
}
\label{tab:reliability_robustness}
\begin{tabular}{lcccc}
\toprule
\multirow{2}{*}{\textbf{Setting}} 
& \textbf{Reliability} 
& \multicolumn{3}{c}{\textbf{Robustness}} \\
\cmidrule(lr){2-2} \cmidrule(lr){3-5}
& \textbf{Repeat Std. $\downarrow$} 
& \textbf{$\rho$ $\uparrow$} 
& \textbf{Mean $\Delta$} 
& \textbf{Std $\Delta$} \\
\midrule
Vague-Prompt & 0.15 & 0.91 & -0.12 & 0.18 \\
Multi-Source & 0.18 & 0.94 & +0.08 & 0.22 \\
Long-Doc     & 0.14 & \textbf{0.96} & +0.05 & 0.15 \\
Multi-Modal  & \textbf{0.11} & 0.93 & -0.03 & \textbf{0.12} \\
\bottomrule
\end{tabular}
\end{table}
\subsection{Ablation Analysis}

\definecolor{rowblue}{RGB}{235,245,255}

\begin{table*}[t]
\centering
\small
\setlength{\tabcolsep}{5.5pt}
\renewcommand{\arraystretch}{1.08}
\caption{
Ablation results of UniPPTAgent on \textsc{Ours-Gemini}. Components are grouped by agent: \textit{Narrative} covers evidence retrieval and visual evidence alignment; \textit{Visual Design} covers layout planning and perceptual refinement. A \checkmark\ indicates the component is enabled. Configurations a--f disable one or more components, while g is the full model. \textit{Overall Avg} is the task-level mean over all settings.
}
\label{tab:ablation_main}
\resizebox{\textwidth}{!}{
\begin{tabular}{c cc cc ccccc}
\toprule
\multirow{2}{*}{Config}
  & \multicolumn{2}{c}{Narrative}
  & \multicolumn{2}{c}{Visual Design}
  & \multicolumn{4}{c}{Setting Avg}
  & \multirow{2}{*}{Overall Avg $\uparrow$} \\
\cmidrule(lr){2-3}\cmidrule(lr){4-5}\cmidrule(lr){6-9}
  & Evid.\ Retr. & Vis.\ Align.
  & Layout & Refine
  & Vague & Long-Doc & Multi-Src & Multi-Modal & \\
\midrule
a & \checkmark &            &            &            & 8.29 & 5.40 & 6.60 & 6.94 & 6.83 \\
b & \checkmark & \checkmark &            &            & 8.05 & 6.19 & 6.86 & 7.75 & 7.29 \\
c &            & \checkmark & \checkmark & \checkmark & 7.70 & 5.25 & 6.41 & 7.92 & 6.97 \\
d & \checkmark &            & \checkmark & \checkmark & 8.29 & 6.43 & 6.92 & 8.09 & 7.52 \\
e & \checkmark & \checkmark &            & \checkmark & 8.23 & 6.54 & 7.12 & 8.10 & 7.58 \\
f & \checkmark & \checkmark & \checkmark &            & 8.14 & 6.76 & 7.02 & 8.07 & 7.58 \\
\midrule
g & \checkmark & \checkmark & \checkmark & \checkmark & \textbf{8.58} & \textbf{6.71} & \textbf{7.17} & \textbf{8.25} & \textbf{7.76} \\
\bottomrule
\end{tabular}
}
\end{table*}

To quantify the contribution of each core component in the UniPPTAgent framework, we conduct ablation experiments on \textsc{Ours-Gemini}. Results are reported in Table~\ref{tab:ablation_main}.

Removing any single component leads to a measurable drop in Overall Avg. \textbf{Evidence retrieval} (config~c) has by far the largest impact, lowering Overall Avg from $7.76$ to $6.97$ ($-0.79$), with the most severe drops in Long-Doc ($-1.46$) and Multi-Source ($-0.76$), confirming that grounded retrieval is the dominant factor for content-heavy tasks. \textbf{Visual evidence alignment}, \textbf{layout planning}, and \textbf{perceptual refinement} (configs~d, e, f) each produce smaller but non-trivial drops ($-0.24$, $-0.18$, $-0.18$), indicating that each component contributes independently. Beyond individual effects, the components are also complementary: retaining only evidence retrieval and visual alignment (config~b) lowers Overall Avg to $7.29$, and further removing visual alignment (config~a) degrades it to $6.83$, with Multi-Modal falling from $8.25$ to $6.94$. These cumulative drops confirm that the four components are jointly necessary.

\section{Conclusion}

Although presentation generation has advanced rapidly, it still lacks a unified formulation and evaluation framework across real-world input settings. Existing studies are usually developed and tested in isolated scenarios, making fair comparison difficult and often obscuring whether systems truly meet the core requirements of grounded presentation generation. To address this gap, we introduce UniPPTBench, a unified benchmark across diverse input settings, and UniPPTEval, an evaluation framework that combines shared metrics, scenario-specific assessment, and transparent reference baselines for reproducible comparison. Experiments show that current systems can often generate visually plausible, presentation-like decks, but remain far less reliable when tasks require strong content grounding, multimodal alignment, or cross-source synthesis. More importantly, good performance on generic presentation-quality metrics does not necessarily imply strong task fulfillment in grounded settings, and system behavior varies markedly across input settings. These findings suggest that presentation generation should be studied not as a single homogeneous task, but as a unified problem with setting-dependent capability demands.

\section*{Acknowledgment}
This work was supported by the CCF-Tencent Rhino-Bird Open Research Fund.

\appendices

\section{Shared Metric Details}
\label{sec:metrics1}

This section provides a detailed description of the construction process for the human-aligned shared metrics, the aggregation strategy used in UniPPTEval, and the implementation-level definitions of the final shared metrics.

\subsection{Shared Metric Construction}

Unlike prior approaches that directly assume a fixed set of evaluation dimensions, UniPPTEval systematically derives a shared metric set from a large-scale candidate metric space through a principled pipeline consisting of ``Selection $\rightarrow$ Validation $\rightarrow$ Orthogonal Combination $\rightarrow$ Efficiency--Accuracy Frontier Selection''.

The candidate metrics are primarily collected from existing works on presentation generation and slide evaluation, including PPTAgent~\cite{zheng2025pptagent}, AutoPresent~\cite{ge2025autopresent}, PresentBench~\cite{chen2026presentbench}, and related systems. In addition, we further expand and refine the metric space based on design literature and real-world user observations. The resulting candidate pool spans multiple complementary dimensions, including content correctness, instruction fulfillment, structural organization, visual quality, and presentation-level effectiveness.

To ensure that automatic evaluation faithfully reflects subjective human preference, we further align candidate metrics with human preference signals. Following recent frameworks on factual evaluation and human alignment analysis~\cite{faitheval,factscore}, we develop a dedicated user study platform and conduct pairwise A/B preference experiments. In each comparison, annotators are asked to select the higher-quality presentation from an anonymized pair. For every candidate metric, we compute the probability that its preference agrees with human judgment to measure its predictive capability.

Experimental results reveal substantial differences among metrics in terms of human alignment. For example, \textit{Visual Integrity} effectively captures users' sensitivity to visual defects and layout errors, demonstrating strong predictive power among individual metrics. However, relying on a single metric remains insufficient for comprehensively characterizing overall presentation quality, since human preference is jointly influenced by multiple aspects including content, structure, visual design, and presentation effectiveness.

Therefore, after retaining metrics with strong human alignment, we further conduct orthogonality analysis to identify redundancy among metrics. Specifically, we perform pairwise correlation analysis over all candidate metrics and preferentially preserve those that simultaneously exhibit strong predictive capability and low mutual correlation. This indicates that these metrics capture complementary quality signals from different perspectives rather than repeatedly measuring the same property.

Building upon this reduced metric space, UniPPTEval adopts a marginal-gain-based greedy combination strategy to progressively construct the shared metric set along the efficiency--accuracy frontier. At each step, the metric providing the largest improvement in prediction accuracy is incorporated, while jointly considering computational cost and evaluation stability. This process ultimately yields a compact shared metric configuration that achieves a favorable balance between evaluation performance and efficiency.

The final shared metrics consist of \textbf{Instruction Fulfillment}, \textbf{Engagement}, \textbf{Content Accuracy}, \textbf{Visual Consistency}, and \textbf{Visual Integrity}. Together, these metrics provide a unified characterization of generated presentations from the perspectives of task completion, communicative effectiveness, factual reliability, and visual quality.

\subsection{Shared Metric Aggregation}

Given the selected shared metrics
$\{m_i\}_{i=1}^{K}$,
UniPPTEval computes the final shared evaluation score through a normalized equal-weight aggregation scheme.

To ensure comparability across heterogeneous evaluation dimensions, each metric is first rescaled into the range $[0,10]$:

\[
\hat{m}_i
=
10 \cdot
\frac{m_i-\min(m_i)}
{\max(m_i)-\min(m_i)+\epsilon},
\]

where $\epsilon$ is a small constant introduced for numerical stability.

The final shared evaluation score is then computed as the arithmetic mean of all normalized shared metrics:

\[
S_{\text{shared}}
=
\frac{1}{K}
\sum_{i=1}^{K}
\hat{m}_i.
\]

We adopt equal-weight aggregation to avoid introducing additional bias toward any individual evaluation dimension. Since the selected metrics are already filtered through human alignment validation and orthogonality analysis, equal aggregation provides a stable and interpretable approximation of overall presentation quality.

\subsection{Shared Metric Definitions}

We next describe the implementation-level definitions of the final shared metrics.
\paragraph{Instruction Fulfillment.} Evaluates whether the generated presentation satisfies explicit and implicit task requirements. It assesses request compliance, target audience, intended tone, slide structure, topic scope, and specified constraints, serving as a macro-level indicator of overall task alignment.

\paragraph{Engagement.} This metric measures the extent to which the presentation is likely to sustain audience attention. It captures narrative flow, framing, slide-to-slide pacing, audience orientation, and the avoidance of monotonous or overly mechanical exposition. This dimension reflects communicative effectiveness, which is essential for real-world presentation quality.

\paragraph{Content Accuracy.} This metric evaluates the factual correctness and internal consistency of the generated presentation. It penalizes hallucinated claims, contradictory information, misleading summaries, and unsupported or implausible statements, providing a comprehensive assessment of content reliability.

\paragraph{Visual Consistency.} This metric assesses whether the presentation maintains a coherent visual system across all slides, including consistency in typography, color palette, layout structure, spacing, and component styles. A presentation may achieve strong local aesthetics but still receive a low score on this metric if its visual language is inconsistent across pages.

\paragraph{Visual Integrity.}
This metric evaluates whether the presentation is free from critical rendering or formatting defects. It focuses on issues such as cropped content, unreadable overlaps, corrupted text, broken layouts, missing images, malformed visual elements, and inconsistencies between visual components and associated textual descriptions.

Specifically, let $N_{\text{error}}$ denote the number of slides containing critical visual defects, and let $N_{\text{total}}$ denote the total number of slides in the presentation. We first compute the defect rate as:

\[
r_{\text{error}}
=
\frac{N_{\text{error}}}
{N_{\text{total}}}.
\]

The final Visual Integrity score is then defined as:

\[
S_{\text{VI}}
=
10 \cdot (1-r_{\text{error}}).
\]

This formulation assigns higher scores to presentations with fewer visually corrupted pages, while explicitly penalizing severe rendering and layout failures that negatively affect readability and usability.

\section{Scenario-Specific Metric Details}
\label{sec:metrics2}
The  scenario-specific metrics are as follows:

\paragraph{\textit{Long-Doc}.}
The long-document setting focuses on source-grounded compression and content selection from extended materials. We define two metrics:

\begin{itemize}

\item \textbf{Key Point Coverage.}
This metric measures whether the generated slides cover the annotated source-grounded content units (e.g., objective facts, arguments, or section-level takeaways). \textbf{Scoring Rule:} The evaluator independently judges the status of each weighted coverage point $i$ as $c_i \in \{0, 0.5, 1\}$, corresponding to not covered, partially covered, and fully covered, respectively. The final score is aggregated using the following weighted formula:
\[
\text{KeyPointCoverage} = \frac{\sum_i w_i \cdot c_i}{\sum_i w_i} \times 10.
\]

\item \textbf{Content Faithfulness.}
Unlike coverage which evaluates recall, faithfulness assesses the factual precision of the generated content. \textbf{Scoring Rule:} The evaluator extracts atomic factual claims from each slide and traces them back to the input source materials for validation~\cite{kryscinski2020evaluating,factscore}. Each claim is assigned a binary state $s_i \in \{0, 1\}$ ($1$ if fully supported by the source document, $0$ if unsupported, contradicted, or fabricated). For $N$ extracted claims, the final score is calculated as:
\[
\text{Faithfulness} = \frac{1}{N}\sum_{i=1}^{N} s_i \times 10.
\]
This reverse-checking mechanism strictly penalizes hallucinations to ensure rigorous evaluation of factual reliability.
\end{itemize}

\paragraph{\textit{Multi-Modal}.}
The multimodal setting evaluates the system's ability to invoke visual evidence (such as charts, illustrations, etc.); its essence is \textit{visual grounding}. For each task, annotations identify critical visual elements and their expected use, including source page references, normalized bounding boxes, paired textual claims, accepted usage modes, and chart-fidelity requirements. We define three metrics:

\begin{itemize}
\item \textbf{Visual Element Utilization.}
This metric measures whether critical visual elements identified in the annotations are materially used in the generated slides. \textbf{Scoring Rule:} Each critical visual element $i$ identified in the annotations is assigned a binary state $v_i \in \{0, 1\}$. Both direct reuse and faithful redraw are considered valid ($v_i = 1$) as long as the core visual information is preserved; otherwise, it is scored as $0$ (omitted). For $N$ annotated visual elements, the final score is calculated as the average utilization rate:
\[
\text{Utilization} = \frac{1}{N}\sum_{i=1}^{N} v_i \times 10.
\]
This metric captures whether the model leverages visual evidence from the input rather than relying solely on textual restatement.
\item \textbf{Figure--Claim Alignment.}
This is a deep visual grounding metric. Drawing on the philosophy of Pathway 2, it verifies whether textual claims are substantively supported by corresponding visual evidence. \textbf{Scoring Rule:} The evaluator assesses every visual element present in the generated slides for its consistency with the textual claims on the same page. Each visual element $i$ is assigned a state $a_i \in \{0, 0.5, 1\}$. A score of $0$ is given if the visual is completely irrelevant, serves merely as a placeholder, or is purely decorative. A score of $0.5$ indicates that the visual is contextually relevant but fails to provide substantive evidentiary support for the claims. A score of $1$ is awarded if the visual perfectly supports the textual claims. For $N$ generated visual elements, the final score is calculated as the average alignment rate:
\[
\text{Alignment} = \frac{1}{N}\sum_{i=1}^{N} a_i \times 10.
\]

\item \textbf{Chart Translation Fidelity.}
This metric measures whether chart information is faithfully preserved when translated into slide-native representations, including key labels, trends, comparisons, numerical relations, and structural semantics. The focus is on preserving chart semantics rather than exact visual replication. This metric captures distortions or hallucinations introduced during chart redrawing or transformation. \textbf{Scoring Rule:} This metric is evaluated exclusively on critical visuals where chart fidelity is explicitly required. Each applicable visual $i$ is assigned a state $f_i \in \{0, 0.5, 1\}$. A score of $1$ is awarded if the original chart is directly inserted and clearly legible, or if it is redrawn while faithfully preserving key values, labels, legends, relationships, and primary takeaways. A score of $0.5$ indicates that the overall structure and main takeaways are maintained, but secondary labels, precision, or auxiliary details are lost. A score of $0$ is given if the visual is missing, replaced solely by text, structurally distorted after redrawing, missing core labels, or if its numerical and comparative relationships contradict the original chart. The final score is calculated as a weighted average over all applicable visuals:
$$\text{Fidelity} = \frac{\sum_i w_i \cdot f_i}{\sum_i w_i} \times 10$$
\end{itemize}

\paragraph{\textit{Multi-Source}.} The multi-source setting evaluates coverage and synthesis across multiple heterogeneous input sources. For each task, annotations specify source-specific contributions, cross-source integration requirements, and overlap groups. These annotations make explicit which information is unique to each source, which content should be synthesized across documents, and which repeated information should be consolidated to avoid redundancy. Based on these annotations, we define three metrics:

\begin{itemize}

\item \textbf{Cross-Document Coverage.}
This metric measures whether the generated slides adequately cover source-specific key information identified in the annotations, rather than disproportionately relying on a single document. It captures whether the model can extract relevant information in a balanced manner from multiple sources. \textbf{Scoring Rule:} The evaluator assesses each annotated source-specific contribution point. Each point $i$ is assigned an effective weight $w_i = \text{source\_weight}_i \times \text{point\_weight}_i$. A state $c_i \in \{0, 0.5, 1\}$ is assigned independently to each point: $1$ if the point is explicitly and substantively preserved; $0.5$ if it is mentioned but lacks completeness or specificity; and $0$ if it is missing or only vaguely alluded to. Annotated accepted signals serve as supporting evidence rather than rigid keyword matching criteria. The final score is calculated as a weighted average:
$$\text{Coverage} = \frac{\sum_i w_i \cdot c_i}{\sum_i w_i} \times 10$$
\item \textbf{Cross-Document Integration.}
This metric evaluates whether information from multiple sources is synthesized into a coherent and unified narrative, rather than simply presenting isolated summaries of each document. \textbf{Scoring Rule:} The evaluator assesses each annotated integration requirement. Each requirement $i$ is assigned a state $s_i \in \{0, 0.5, 1\}$. A score of $1$ is awarded if the slides explicitly synthesize the relevant sources into a comparison, bridge, trade-off, joint framework, or unified storyline. A score of $0.5$ indicates shallow, incomplete, or unilateral integration. A score of $0$ is given if the sources are only mentioned separately without genuine synthesis (i.e., parallel summarization does not count as integration). The final score is calculated as a weighted average:
$$\text{Integration} = \frac{\sum_i w_i \cdot s_i}{\sum_i w_i} \times 10$$
\item \textbf{Redundancy Deduplication.}
This metric measures whether redundant themes across multiple sources are consolidated into a single coherent expression, rather than being repeatedly stated source-by-source. \textbf{Scoring Rule:} The evaluation is based on annotated overlap groups (themes prone to repetition). The evaluator follows a two-step process. First, a group is deemed applicable only if its underlying theme is actually present in the generated slides. Second, each applicable group $i$ is assigned a state $d_i \in \{0, 0.5, 1\}$. A score of $1$ is awarded if the overlapping material is successfully consolidated into a unified treatment. A score of $0.5$ indicates partial consolidation with some residual stitched or redundant repetition. A score of $0$ is given if the theme is repeatedly presented without effective consolidation. The final score is calculated as a weighted average exclusively over the applicable groups:
\[
\text{Deduplication} = \frac{\sum_i w_i \cdot d_i}{\sum_i w_i} \times 10.
\]
\end{itemize}

\paragraph{\textit{Vague-Prompt}.} The vague-prompt setting does not introduce scenario-specific metrics in the current version because it lacks stable source-grounded references. These tasks are therefore evaluated only with shared metrics, avoiding hidden assumptions about what source-specific content should have appeared in the generated deck.

\section{Implementation Details of UniPPTAgent Baselines}
\label{app:baseline_impl}

\paragraph{Document parsing and knowledge construction.}
Reference documents are parsed via the MinerU cloud API, which extracts body text, headings, tables, figures, and associated captions. Each parsed document is organized into three granularities of knowledge following narrative agent: a condensed \textit{card} using the first 1{,}200 characters of the abstract for narrative planning, section-level \textit{facts} with up to 1{,}800 characters per section for outline induction, and fine-grained \textit{chunks} constructed with a 900-character sliding window and 200-character overlap for evidence retrieval. These representations are merged into the unified knowledge base $K$.

\paragraph{Progressive content structuring.}
The Narrative Agent uses DeepSeek-V3 for all three stages of the plan--outline--ground procedure. Narrative planning determines the global structure and slide count from $K_{\text{card}}$; outline induction expands the plan into page-level titles, key messages, and content points guided by $K_{\text{facts}}$; and evidence retrieval scores passages in $K_{\text{chunks}}$ by cosine similarity against each page's outline entry using Qwen \texttt{text-embedding-v3}. The system greedily retains the top-ranked passages up to 12 passages or 2{,}000 characters, whichever comes first. The resulting outline $O$ and retrieved evidence $G$ are passed to the variant backbone to synthesize page descriptions $D=\{d_i\}$.

\paragraph{Visual evidence alignment.}
For documents containing visual assets, each candidate figure is matched to the slide page whose content it best supports. We use a coarse-to-fine strategy. In the coarse stage, the caption and surrounding context of each extracted figure are embedded together with each page description using Qwen \texttt{text-embedding-v3}, and pairs with cosine similarity below 0.30 are discarded. In the fine stage, a VLM, \texttt{gemini-3-flash-preview}, re-ranks the remaining candidates by jointly assessing visual and textual relevance. At most 5 figures are retained for each page and appended to the corresponding page description.

\paragraph{Style induction and HTML generation.}
The Style Agent uses the variant backbone to perform schema-guided slot filling over the Style Schema, producing a global style $S_{\text{global}}$ decomposed into opening, body, and ending modules. In the Visual Design Agent, the variant backbone first generates an abstract layout blueprint for each page, specifying element types and bounding boxes in $[0,1]$-normalized coordinates relative to a 16:9 canvas. Conditioned on this blueprint and the global style, the backbone generates structured HTML with temperature 0.4 and a maximum of 8{,}192 output tokens per page. If the generated HTML fails structural validation, such as malformed markup or missing required elements, the page is retried up to 3 times.

\paragraph{Perceptual refinement.}
Each rendered page undergoes iterative diagnosis and repair. In each iteration, the page HTML is rendered with a headless Playwright browser, and the resulting screenshot is inspected by a DOM-level overflow detector and a VLM, \texttt{gemini-3-flash-preview}. The detector flags four categories of visual defects: \texttt{OVERFLOW\_CROPPED}, \texttt{OVERLAP\_UNREADABLE}, \texttt{GARBLED\_RENDERING}, and \texttt{IMAGE\_TEXT\_MISMATCH}. Detected errors, together with the screenshot and current HTML, are fed back to the variant backbone, which generates a targeted HTML patch. The loop terminates after at most 5 iterations or when no errors remain.

\bibliographystyle{IEEEtran}

\bibliography{references}

\section{Biography Section}
\vspace{-2em}

\begin{IEEEbiographynophoto}{Bo Zhao}
is currently pursuing the doctor degree in Shanghai Innovation Institute. His current research interests include multimodal large language models and content creation.
\end{IEEEbiographynophoto}

\vspace{-2.0em}

\begin{IEEEbiographynophoto}{Maosheng Pang}
is currently pursuing the M.S. degree in Artificial Intelligence at the School of Intelligence Science and Technology, Nanjing University, Nanjing, China. His research interests include AI Agent and generative retrieval.
\end{IEEEbiographynophoto}
\vspace{-2.0em}

\begin{IEEEbiographynophoto}{Huan Yang}
 received his B.S. degree in computer science in 2014 and Ph.D. degree in computer science in 2019, both from Shanghai Jiao Tong University. He is currently a senior algorithm expert at Kuaishou Kling Group, leading the Kolors base model team. Previously, he served as a senior algorithm expert at 01.AI's multi-modality group, leading the multi-modal generation direction, and was a senior researcher at Microsoft Research Asia's multi-modal computing group. His research interests primarily include GAN/diffusion-based content generation (image/video generation and manipulation) and low-level vision (image/video restoration and enhancement). He has published approximately 20 papers at top-tier international conferences, including CVPR, ICCV, ECCV, and NeurIPS.
\end{IEEEbiographynophoto}
\vspace{-1.2em}

\begin{IEEEbiographynophoto}{Chen Zhang}
is currently pursuing the B.S. degree with the School of Management, Nanjing University, Nanjing, China. Her current research interests include multimodal large language models. She was a recipient of the First-Class People's Scholarship.
\end{IEEEbiographynophoto}
\vspace{-1.7em}

\begin{IEEEbiographynophoto}{Yixin Cao}
is a professor at Institute of Trustworthy Embodied Al, Fudan University. He is also obtained his Ph.D. from Tsinghua University and has held positions as a research fellow, research assistant professor, and assistant professor at the National University of Singapore, Nanyang Technological University, and Singapore Management University, respectively. His research areas include natural language processing, knowledge engineering, and multimodal alignment. He has published over 100 papers at international renowned conferences and journals, with more than 12000 citations on Google. His research achievements have been awarded the Best Paper/Nomination at two international conferences. He has received Lee Kong Chian Fellowship, Google South Asia \& Southeast Asia Awards, the AI2000 Most Influential Scholar honorable mention, and Top 2\% Global Scientists of 2024, 2025 by Elsevier. Cao Yixin also serves as the demonstration program chair at ACL2024, local chair at WSDM2023, finance chair at the Web Conference2024, and area chair for multiple international conferences.
\end{IEEEbiographynophoto}
\vspace{-1.2em}

\begin{IEEEbiographynophoto}{Wei Ji}
is an associate professor at Nanjing University.
He was a Research Fellow in the School of Computing at National University of Singapore. He received the Ph.D. degree from Zhejiang University. He has published more than 80 papers in top-tier venues such as CVPR,
ECCV, SIGIR, AAAI, TPAMI, TIP, and TCYB. He serves as Associate Editor of IEEE TIP, IEEE TCSVT, ACM TOMM and Area Chair of NeurIPS, IJCAI, ACM Multimedia, and so on. His current research interests include multi-modal learning, vision and language, and cross-modal retrieval.
\end{IEEEbiographynophoto}

\vfill

\end{document}